\documentclass{article}
\usepackage{spconf,amsmath,graphicx,hyperref}

\usepackage{latexsym}
\usepackage{amssymb}
\usepackage{amsmath}
\usepackage{amsthm}
\usepackage{booktabs}
\usepackage{enumitem}
\usepackage{graphicx}
\usepackage{color}
\usepackage{algorithm}
\usepackage{algorithmic}
\usepackage{multirow}

\newtheorem{theorem}{Theorem}
\newtheorem{lemma}{Lemma}

\newtheorem{proposition}{Proposition}

\newtheorem{definition}{Definition}
\newtheorem{assumption}{Assumption}

\title{Pretrain-DPFL: Mitigating Noise Detriment in \\
Differentially Private Federated Learning with Model Pre-training}

\name{
Huitong Jin$^{1}$ \qquad
Yipeng Zhou$^{2}$ \qquad
Quan Z. Sheng$^{2}$ \qquad
Shiting Wen$^{3}$ \qquad
Laizhong Cui$^{1\star}$
\thanks{$^{\star}$ Corresponding author: cuilz@szu.edu.cn}
\thanks{This work has been partially supported by National Natural Science Foundation of China under Grant No. U23B2026 and No.62372305, Guangdong Basic and Applied Basic Research Foundation under Grant No. 2024B1515040012, Shenzhen Science and Technology Program under Grant No. KJZD20230923114809020, and Research Team Cultivation Program of Shenzhen University, Grant No.2023QNT015 and Key Programs of Ningbo Municipal Natural Science Foundationn Under Grant No. 2024J021.}
}
\address{$^{1}$ Shenzhen University, China \quad $^{2}$ Macquarie University, Australia \quad $^{3}$ NingboTech University, China}

\begin{document}

%\ninept
%
\maketitle
\begin{abstract}
Differentially Private Federated Learning (DPFL) strengthens privacy protection by perturbing model gradients with noise, though at the cost of reduced accuracy. Although prior empirical studies indicate that initializing from pre-trained rather than random parameters can alleviate noise disturbance, the problem of optimally fine-tuning pre-trained models in DPFL remains unaddressed. In this paper, we propose Pretrain-DPFL, a framework that systematically evaluates three most representative fine-tuning strategies: full-tuning (FT), head-tuning (HT), and unified-tuning(UT) combining HT followed by FT. Through convergence analysis under smooth non-convex loss, we establish theoretical conditions for identifying the optimal fine-tuning strategy in Pretrain-DPFL, thereby maximizing the benefits of pre-trained models in mitigating noise disturbance. Extensive experiments across multiple datasets demonstrate Pretrain-DPFL's superiority, achieving $25.22\%$ higher accuracy than scratch training and outperforming the second-best baseline by $8.19\%$, significantly improving the privacy-utility trade-off in DPFL.
\end{abstract}
\section{Introduction}
\label{sec:intro}
Differentially Private Federated Learning (DPFL) is the de facto standard privacy preservation framework for model training across multiple clients \cite{ZhengCoGAP, Tsouvalas2024}. DPFL improves Federated Learning (FL) by adding Gaussian \cite{abadi2016deep} or Laplace \cite{wei2020federated} noise to gradients to protect privacy, though at the cost of model performance degradation~\cite{tramer2021differentially, Francis2025Differentially}. Previous efforts have been dedicated to mitigating Differentially Private (DP) noise influence by developing specialized architectures for private training \cite{yang2023privatefl, ShahmiriLaplace}, or reducing model size during training \cite{ma2023optimized, cui2022boosting}.

% In FL, multiple rounds of training are conducted, during which a coordinating server periodically exchanges gradients with clients \cite{mcmahan2017communication}. However, these original gradients inherently carry the risk of leaking sensitive information, making them unsafe and susceptible to privacy breaches \cite{geiping2020inverting, liu2024please, zarifzadeh2024low}. DPFL introduces zero-mean noises from Gaussian \cite{abadi2016deep} or Laplace \cite{wei2020federated} distributions to obfuscate the gradients to be exposed for enhancing privacy protection.

% The addition of differentially private (DP) noise can greatly complicate the model training process, and probably significantly compromise model performance \cite{tramer2021differentially, ma2023optimized}. Previous efforts have been dedicated to mitigating DP noise influence by developing specialized architectures for private training \cite{tramer2021differentially, yang2023privatefl}, or reducing the model size during training \cite{cui2022boosting, ma2023optimized}. Orthogonal to existing methods, DP fine-tuning approaches leveraging pre-training models to enhance DPFL performance have recently garnered increasing attention.

Orthogonal to existing methods, DP fine-tuning approaches leveraging pre-training models to boost DPFL have recently garnered increasing attention\cite{li2022does, bao2025unlocking}.
DP fine-tuning typically begins with a pre-trained model obtained from training on public datasets (e.g., ImageNet-1K) without privacy constraints, followed by private fine-tuning on specific downstream datasets (e.g., CIFAR-10) \cite{ganesh2023public}. Recent studies reported that the use of pre-trained parameters in DPFL can accelerate model convergence, reduce model exposure times, and hence mitigate noise influence, compared to scratch training (ST) which begins with randomly initialized parameters
\cite{li2022does, de2022unlocking, wang2024neural, ke2024convergence}. In particular, \cite{de2022unlocking} has empirically tried different fine-tuning  strategies, such as head-tuning (HT) and full-tuning (FT). \cite{ke2024convergence} proposed  unified-tuning (UT)  that first HT followed by FT and theoretically derived the performance gaps between HT, FT, and UT under varying noise scales. However, these studies primarily focus on Machine Learning (ML) scenarios without specifying the optimal fine-tuning strategy under different conditions. In particular, \cite{ke2024convergence} explored the optimal fine-tuning strategy by simultaneously attempting different strategies. This approach is impractical in DPFL, as it would lead to a multiplicative privacy budget consumption due to the exposure of gradients in each attempt.

To identify the optimal fine-tuning strategy in DPFL with pre-trained models, we introduce the Pretrain-DPFL framework, one of the first approaches capable of automatically selecting the optimal strategy.
From a learning perspective, FT is expected to outperform HT, as HT updates only a small portion of model parameters (i.e., a few layers). From a privacy perspective, however, HT incurs significantly lower privacy loss than FT, since only the fine-tuned parameters are exposed. Balancing these competing factors makes it challenging to determine the optimal strategy in practical DPFL.
To address this challenge, we formalize the UT strategy that conducts $T_1$ rounds of HT followed by $T-T_1$ rounds of FT, where $T$ is the total number of model exposure rounds.
We analyze the convergence of this unified strategy under smooth non-convex loss, through which we can quantify the performance gap between FT and HT under different conditions, and further determine the best strategy with the optimal $T_1$.

% In summary, our contributions are three-fold. \emph{Firstly}, we demonstrate that pre-trained parameters can significantly mitigate noise influence and enhance DPFL performance. \emph{Secondly},  we propose the Pretrain-DPFL framework. Based on convergence analysis, we can determine the optimal fine-tuning strategy for DPFL with pre-trained models. \emph{Lastly}, extensive experimental results demonstrate that Pretrain-DPFL significantly outperforms existing methods by improving $21.2\%$ accuracy over ST and $5.91\%$ accuracy over the second-best baseline under a fixed privacy budget. 

In summary, our contributions are three-fold. \emph{Firstly}, we analyze the convergence of HT, FT, and UT under smooth non-convex loss and derive conditions for selecting the optimal strategy. \emph{Secondly}, we propose the Pretrain-DPFL framework, which automatically selects the optimal fine-tuning strategy for DPFL without incurring additional privacy cost from trial-and-error tuning. \emph{Lastly}, extensive experiments on multiple datasets show that Pretrain-DPFL improves accuracy by 25.22\% over ST and 8.19\% over the second-best baseline under a fixed privacy budget. 

\section{Preliminaries}
% In this section, we introduce preliminary knowledge on DP and DPFL with different fine-tuning strategies.

\subsection{Differential Privacy}
% Differential Privacy (DP) is a rigorous privacy-preserving framework originally applied in databases \cite{ma2023optimized}.
 %We review the DP definition \cite{dwork2014algorithmic, ma2023optimized} as follows.

\begin{definition}
\label{def: dp}
\textbf{\((\epsilon, \delta)\)-Differential Privacy \cite{Francis2025Differentially}:} A randomized algorithm \(\mathcal{M}\) satisfies \((\epsilon, \delta)\)-DP if for any two datasets \( \mathcal{D}\) and \(\mathcal{D}'\) differing by one element, and for any subset \(\mathcal{S}\) of outputs:
\begin{equation}
\Pr[\mathcal{M}(\mathcal{D}) \in \mathcal{S}] \leq \exp(\epsilon) \Pr[\mathcal{M}(\mathcal{D}') \in \mathcal{S}] + \delta,
\label{eqs: dpdefinition}
\end{equation}
\end{definition}

Here, \( \epsilon \) is the privacy budget and \( \delta \) is the failure probability; smaller values indicate stronger privacy.

\subsection{DPFL with Different Fine-tuning Strategies}
We adopt FedSGD as the DPFL framework\cite{mcmahan2017communication, zhou2023optimizing} with $N$ clients jointly training global model $F(\theta)$,  where $\theta \in \mathbb{R}^{p+n}$ includes with feature extractor($\mathbb{R}^p$) and classifier head ($\mathbb{R}^n$). we formally define the three fine-tuning strategies:

\begin{definition}
\label{def: ht}
    \textbf{Head-Tuning Strategy (HT):} Only the last $n$ parameters are trainable. After local updates and aggregation, the global update at iteration $t$ is: $\tilde{\theta}_{t+1} = \sum_{i=1}^{N}\frac{d_i}{d} [\tilde{\theta}_t - \eta_t (\mathbf{m} \odot \mathbf{g}_t^i + \tilde{\mathbf{w}}_t^i)] = \tilde{\theta}_t - \eta_t (\mathbf{m} \odot \mathbf{g}_t + \tilde{\mathbf{w}}_t)$. 
    
    Here, $\eta_t$ is the learning rate, $\mathbf{m} = [0_p, 1_n]^T$ masks the gradient, $\mathbf{g}_t = \sum_{i=1}^N \frac{d_i}{d} \mathbf{g}_t^i$ is the aggregated gradient, and $\tilde{\mathbf{w}}_t = \sum_{i=1}^N \frac{d_i}{d} \tilde{\mathbf{w}}_t^i$ represents the aggregated DP noise.
\end{definition}

\begin{definition}
\label{def: ft}
    \textbf{Full-Tuning Strategy (FT):} All $p+n$ parameters are trainable. After local updates and aggregation, the global update at iteration $t$ is: $\hat{\theta}_{t+1} = \sum_{i=1}^{N}\frac{d_i}{d} [\hat{\theta}_t - \eta_t (\mathbf{g}_t^i + \hat{\mathbf{w}}_t^i)] = \hat{\theta}_t - \eta_t (\mathbf{g}_t + \hat{\mathbf{w}}_t)$.
    
    Here, $\hat{\mathbf{w}}_t^i \in \mathbb{R}^{p+n}$ represents the DP noise added by client $i$, and $\hat{\mathbf{w}}_t = \sum_{i=1}^N \frac{d_i}{d} \hat{\mathbf{w}}_t^i$ denotes the aggregated noise.
    \end{definition}

\begin{definition}
\label{def: ht-ft}
    \textbf{Unified-Tuning Strategy (UT):} Suppose that DPFL will conduct $T$ global iterations. The UT strategy consists of two phases: the first $T_1$ ($0 < T_1 < T$) iterations of HT, followed by $T - T_1$ iterations of FT.
\end{definition}

In our analysis, we will study how to set $T_1$ for determining the optimal fine-tuning strategy.

\begin{algorithm}[ht]
\caption{Pretrain-DPFL FrameWork}
\label{alg:Pretrain-DPFL}
\begin{algorithmic}[1]
\STATE \textbf{Step 1:} Pretrain on ImageNet-1K to obtain $\theta_0$.
\STATE \textbf{Step 2:} Estimate $G_1^2,G_2^2,\Lambda_1^2,\Lambda_2^2,L,\Gamma$ from clients.
\STATE \textbf{Step 3:} Compute $T_1$ based on Theorem \ref{theorem: laplace-convergence} or \ref{theorem: gaussian-convergence}.

\STATE \textbf{Step 4: Main training}  
\FOR{ each round $t=0,\dots,T-1$ }  
\FOR{each client $i$}
\STATE Sample batch $\mathcal{B}_t^i$ from $\mathcal{D}_i$.
\STATE Generate $\tilde{\mathbf{w}}_t^i$ (or $\hat{\mathbf{w}}_t^i$) accoding to DP mechanism.
\STATE Update parameters as follows
\[
\theta_{t+1}^i =
\begin{cases}
\theta_t^i - \eta_t(\mathbf{m}\odot\nabla F_i + \tilde{\mathbf{w}}_t^i), & t \le T_1 ~ \text{(HT)} \\[6pt]
\theta_t^i - \eta_t(\nabla F_i + \hat{\mathbf{w}}_t^i), & t > T_1 ~ \text{(FT)}
\end{cases}
\]\ENDFOR
\STATE Server aggregates: $\theta_{t+1} = \sum_{i=1}^N \tfrac{d_i}{d}\theta_{t+1}^i$
\ENDFOR
\end{algorithmic}
\end{algorithm}

%%%%%%%%%%%%%%%%%%%%%%%%%%%%%%%%%%%%%%%%%%%%%%%%%%%%%%%%%%%%%%%%%%%%%%%%

\section{Pretrain-DPFL Framework}
\subsection{Pretrain-DPFL}
The DPFL using pre-trained parameters (Pretrain-DPFL) framework is presented in Algorithm \ref{alg:Pretrain-DPFL}. Firstly, model parameters are pretrained on a public dataset (e.g., ImageNet-1K \cite{imagenet}) in a centralized manner on the server, yielding the initialization $\theta_0$. Secondly, each client locally estimates $G_1^2, G_2^2, \Lambda_1^2, \Lambda_2^2, L$, and $\Gamma$ using the algorithm proposed in \cite{zhou2023optimizing, wang2019adaptive, chen2020convergence}, and transmits the results to the server, which then aggregates them to obtain the final global estimates. Based on the aggregated values, the server computes $T_1$ according to Theorem~\ref{theorem: laplace-convergence} or Theorem~\ref{theorem: gaussian-convergence}. Finally, the training proceeds for $T$ global rounds: each client samples a minibatch, adds DP noise accoding to DP mechanism, and updates local parameters under either HT or FT strategy depending on $t \le T_1$ or $t > T_1$. The server aggregates the updates to obtain the global model. In this way, Pretrain-DPFL adaptively selects the optimal fine-tuning strategy, ensuring that the model achieves the best utility-privacy performance.

Specifically, Step 2 incurs only minimal overhead: each client requires few iterations (set to 5 in our experiments) to obtain the statistics, leading to an additional time cost of less than 10\%. Moreover, only 6 scalar values are transmitted to the server, resulting in $\mathcal{O}(1)$ communication cost. Compared with sharing gradients or model weights, these statistics reveal negligible private information. For additional safety, we allocate a privacy budget of 0.01 to protect the reported statistics, thereby further strengthening privacy guarantees.

% The DPFL using pre-trained parameters (Pretrain-DPFL) framework is presented in Algorithm \ref{alg:Pretrain-DPFL}. Pretrain-DPFL first utilizes a public dataset (e.g., ImageNet-1K \cite{imagenet}) to pretrain the model parameters in a centralized manner on the server without privacy concerns. The pre-trained parameters are then used as the initialized model. Subsequently, the \emph{StrategySelection} algorithm is invoked to determine the fine-tuning strategy, which will be detailed in the next subsection. Pretrain-DPFL uses the optimal fine-tuning strategy to perform a total of $T$ global iterations.

\subsection{Theoretical Analysis}
In DPFL, the injected noise $\tilde{\mathbf{w}}_t^i$ and $\hat{\mathbf{w}}_t^i$ follow Laplace or Gaussian mechanisms. We briefly summarize key definitions and assumptions before presenting the convergence theorems.

\begin{theorem}
\label{theorem:laplace}
Laplace Mechanism \cite{wei2020federated}. For a query with the $l_1$-sensitivity, assuming that gradients are $l_1$-bounded by $\xi_1$, the Laplace mechanism ensures $(\epsilon_i, 0 )$-DP by adding $\mathbf{\tilde{w}}_t^i \sim \text{Lap}\left(0, \frac{2T\xi_1}{d_i \epsilon_i} \mathbb{I}_{n}\right)$, $\mathbf{\hat{w}}_t^i \sim \text{Lap}\left(0, \frac{2T\xi_1}{d_i \epsilon_i} \mathbb{I}_{p+n}\right)$.
\end{theorem}

% \begin{theorem}
% \label{theorem:gaussian}
% Gaussian Mechanism \cite{abadi2016deep}. For gradients bounded in $l_2$-norm by $\xi_2$, Gaussian mechanism ensures $(\epsilon_i, \delta_i)$-DP by adding $\mathbf{\tilde{w}}_t^i \sim \mathcal{N}(0, \sigma_i^2 \mathbb{I}_{n})$ and $\mathbf{\hat{w}}_t^i \sim \mathcal{N}(0, \sigma_i^2 \mathbb{I}_{p+n})$, with $\sigma_i^2 \ge \frac{c_2^2 \xi_2^2T}{d_i^2 \epsilon_i^2}\log \frac{1}{\delta_i}$.
% \end{theorem}

\begin{proposition}
\label{prop:laplace variance}
    Variance of aggregated Laplace noise: The expected $L_2$-norm squared of the aggregated Laplace noise over $N$ clients is: $\mathbb{E}[\|\tilde{\mathbf{w}}_t\|_2^2] = \frac{8nT^2\xi_1^2}{d^2} \sum_{i=1}^N \frac{1}{\epsilon_i^2}$ for HT or $\mathbb{E}[\|\hat{\mathbf{w}}_t\|_2^2] = \frac{8(p + n)T^2\xi_1^2}{d^2} \sum_{i=1}^N \frac{1}{\epsilon_i^2}$ for FT.
\end{proposition}

% \begin{proposition}
% \label{prop:gaussian variance}
%     Variance of aggregated Gaussian noises: The expected $L_2$-norm squared of the aggregated Gaussian noise over $N$ clients is: $\mathbb{E} [\| \tilde{\mathbf{w}}_t \|_2^2] = \frac{c_2^2nT\xi_2^2}{d^2} \sum_{i=1}^{N} \frac{1}{\epsilon_i^2} \log \frac{1}{\delta_i}$ and $\mathbb{E} [\| \hat{\mathbf{w}}_t \|_2^2] = \frac{c_2^2(p+n)T\xi_2^2}{d^2} \sum_{i=1}^{N} \frac{1}{\epsilon_i^2} \log \frac{1}{\delta_i}$.
% \end{proposition}

 % {\bf Remarks I.} TODO 把这个放到最后的总结讨论那里
 
 % The variance of FT noise is $\tfrac{p+n}{n}$ times larger than that of HT (e.g., $\sim 422$ for CNNs \cite{ma2023optimized}). Despite higher noise, FT enables stronger learning by tuning all parameters, creating a trade-off in strategy selection.

\begin{definition}
\label{def: non-iid degree}
The Bound of Non-IID Degree \cite{ma2023optimized}: For any model $\theta$, we define $\Gamma \ge \mathbb{E} [\| \nabla F_i(\theta) - \nabla F(\theta) \|_2^2]$.
\end{definition}

% We simplify the participation scheme by following the assumption in \cite{Li2020On} that assigns an equal weight (\(d_1 = d_2 = \dots = d_N\)) to all clients. Additional assumptions from previous works \cite{basu2019qsparse, ma2023optimized} are introduced to further streamline the analysis, as outlined below.

\begin{assumption}
\label{asm: l-smooth}
    $L$-smoothness \cite{basu2019qsparse}: All local loss functions $F_i$ are $L$-smooth, i.e., for any $\theta, \theta' \in \mathbb{R}^{p+n}$, we have $F_i (\theta) \leq F_i(\theta ') + \langle \nabla F_i(\theta'), \theta - \theta' \rangle + \frac{L}{2} \| \theta - \theta' \|_2^2$, for $i = 1, \dots, N$.
\end{assumption}

% \begin{assumption}
% \label{asm: bounding-variances}
%     Bounded variances and second momentum: There exist constants $\Lambda_2 >0, \Lambda_1 > 0$ and $G_1 \ge \Lambda_1, G_2 \ge \Lambda_2$ such that: $\mathbb{E}_{\mathcal{B}_t^i \subset \mathcal{D}_i} [\|\nabla F_i(\tilde{\theta}_{t-1}; \mathcal{B}_t^i) - \nabla F_i(\tilde{\theta}_{t-1}) \|_2^2]  \leq \Lambda_1^2,$ $\mathbb{E}_{\mathcal{B}_t^i \subset \mathcal{D}_i} [\|\nabla F_i(\hat{\theta}_{t-1}; \mathcal{B}_t^i) - \nabla F_i(\hat{\theta}_{t-1}) \|_2^2]  \leq \Lambda_2^2,$ and $\mathbb{E}_{\mathcal{B}_t^i \subset \mathcal{D}_i} [\|\nabla F_i(\tilde{\theta}_{t-1}; \mathcal{B}_t^i)\|_2^2]  \leq G_1^2,$ $\mathbb{E}_{\mathcal{B}_t^i \in \mathcal{D}_i} [\|\nabla F_i(\hat{\theta}_{t-1}; \mathcal{B}_t^i)\|_2^2]  \leq G_2^2.$
% \end{assumption}

\begin{assumption}
\label{asm: bounding-variances}
Bounded variance and second moment \cite{ma2023optimized}:
For any $\theta \in \mathbb{R}^n$ with $t \ge 1$, there exist constants $\Lambda_1,\Lambda_2>0$ and $G_1 \ge \Lambda_1,\, G_2 \ge \Lambda_2$ such that $\mathbb{E}_{\mathcal{B}_t^i \subset \mathcal{D}_i} [\|\nabla F_i(\theta; \mathcal{B}_t^i) - \nabla F_i(\theta) \|_2^2]  \leq \Lambda_j^2$ and $\mathbb{E}_{\mathcal{B}_t^i \subset \mathcal{D}_i} [\|\nabla F_i(\theta; \mathcal{B}_t^i)\|_2^2]  \leq G_j^2$, for $\theta=\tilde{\theta}_{t-1}$ ($j=1$) and $\theta=\hat{\theta}_{t-1}$ ($j=2$).
\end{assumption}

Let $v_t$ denote FT parameters after $t$ rounds without noise. Based on the Assumption \ref{asm: bounding-variances} and Proposition \ref{prop:laplace variance}, we derive:

\begin{lemma}
\label{lemma:laplace-bounding-ht-distance}
    Bounding the Distance Between Noisy HT and Noise-Free FT Parameters under Laplace Mechanism:
    \begin{equation}
        \mathbb{E} [\| v_t - \tilde{\theta}_t \|_2^2] \leq 2\eta^2[2(G_1^2 + G_2^2) + \frac{8n T^2 \xi_1^2}{d^2} \sum_{i=1}^{N} \frac{1}{\epsilon_i^2}].
    \end{equation}
\end{lemma}

\begin{lemma}
\label{lemma:laplace-bounding-ft-distance}
    Bounding the Distance Between Noisy and Noise-Free FT Parameters under Laplace Mechanism:
    \begin{equation}
        \mathbb{E} [\| v_t - \hat{\theta}_t \|_2^2] \leq \eta^2(\frac{8(n+p) T^2 \xi_1^2}{d^2} \sum_{i=1}^{N} \frac{1}{\epsilon_i^2}).
    \end{equation}
\end{lemma}

% In Lemma \ref{lemma:laplace-bounding-ht-distance}, the first term $2(G_1^2 + G_2^2)$ approximates the difference in learning capability between HT and FT during training, while the second term represents the noise variance. Comparing this with Lemma \ref{lemma:laplace-bounding-ft-distance}, we observe that FT introduces larger aggregation noise variance in a single training iteration, whereas HT leads to a gap in learning capability. This reveals a delicate trade-off between learning and privacy which actually can be tuned by $T_1$.

% In Lemma \ref{lemma:laplace-bounding-ht-distance}, $2(G_1^2+G_2^2)$ reflects the learning gap between HT and FT, while the second term denotes noise variance. Compared with Lemma \ref{lemma:laplace-bounding-ft-distance}, FT incurs higher noise, whereas HT suffers from weaker learning capacity, revealing a trade-off between learning and privacy tunable via $T_1$.

% The proof of Lemmas \ref{lemma:laplace-bounding-ht-distance} and \ref{lemma:laplace-bounding-ft-distance} are presented in  Appendix. Based on these lemmas, we further prove the main theorem as follows. 

\begin{theorem}
\label{theorem: laplace-convergence}
    Let Assumptions \ref{asm: l-smooth} and \ref{asm: bounding-variances} hold. Let the learning rate $\eta = \frac{\beta}{\sqrt{T}}$ where $\beta < \frac{\sqrt{T}}{2L}$. The convergence rate of Pretrain-DPFL under the Laplace Mechanism is as follows:
    \begin{equation}
        \label{eqs: laplace-convergence}
        \begin{aligned}
            & \quad \mathbb{E}[\| \nabla F(\mathbf{x}_T) \|_2^2] \\
            &= \frac{1}{T} (\sum_{t=0}^{T_1} \mathbb E[\| \nabla F (\tilde{\theta}_t)\|_2^2] + \sum_{t=T_1 + 1}^{T} \mathbb{E} [\| \nabla F(\hat{\theta}_t) \|_2^2]) \\
            & \leq C_0 + C_1T_1 + C_2(T-T_1),
        \end{aligned}
    \end{equation}
    where $C_0 = \frac{4\{\mathbb{E}[F(\theta_{0})] - F^*\}}{\beta T^{1/2}} + (\frac{64\beta^2 L^2 n \xi_1^2}{d^2} \sum_{i=1}^{N} \frac{1}{\epsilon_i^2}) T$, $C_1 = \frac{4\beta}{T^{3/2}} (\frac{L\Lambda_1^2}{N} + 2L\Gamma) + \frac{16\beta^2L^2}{T^2} (G_1^2 + G_2^2)$, and $C_2 = \frac{4\beta}{T^{3/2}}(\frac{L\Lambda_2^2}{N} + 2L\Gamma) + \frac{64 \beta^2 L^2 p \xi_1^2}{d^2} \sum_{i=1}^{N} \frac{1}{\epsilon_i^2}$.
\end{theorem}

We sketch the proof of Theorem \ref{theorem: laplace-convergence} as follows. Under the given assumptions, we first obtain $\mathbb{E}[F(v_{t+1})] - F(v_t) \leq -\tfrac{\eta}{4}\|\nabla F(\tilde{\theta}_t)\|_2^2 + \eta L^2 \| v_t - \tilde{\theta}_t \|_2^2 + 2L\eta^2\Gamma + \tfrac{L\eta^2 \Lambda_1^2}{N}$. 
By applying Lemma \ref{lemma:laplace-bounding-ht-distance} and taking expectations, we can bound $\mathbb{E}[|\nabla F(\tilde{\theta}_t)|_2^2]$. Similarly, invoking Lemma \ref{lemma:laplace-bounding-ft-distance} allows us to bound $\mathbb{E}[|\nabla F(\hat{\theta}_t)|_2^2]$. Summing over all iterations and simplifying yields the desired result for $\mathbb{E}[|\nabla F(\mathbf{x}_T)|_2^2]$.

The main insights from Theorem \ref{theorem: laplace-convergence} include: 
(1) In the absence of noise, all terms vanish as $T \to \infty$, ensuring convergence.
(2) By ignoring $\Lambda_1$ and $\Lambda_2$, the magnitudes of $C_1$ and $C_2$ are primarily determined by $\frac{16\beta^2L^2}{T^2}(G_1^2 + G_2^2)$ and $\frac{64 \beta^2 L^2 p \xi_1^2}{d^2} \sum_{i=1}^{N} \frac{1}{\epsilon_i^2}$, respectively.\footnote{$\Lambda_1$ and $\Lambda_2$ can be ignored because $\Lambda_1\ll G_1$ and $\Lambda_2\ll G_2$, which can be verified by experients.} The former represents the learning capability gap between HT and FT, while the latter reflects the additional noise impact of FT by tuning $p$ more parameters.
(3) For fixed $\epsilon$ and $T$, the optimal $T_1$ minimizing $\mathbb{E}[|\nabla F(\mathbf{x}_T)|_2^2]$ can be identified: if $C_1 \geq C_2$, the optimal configuration is $T_1 = T$, indicating that \emph{the HT fine-tuning strategy should be selected}. Conversely, when $C_1 < C_2$, the optimal choice is $T_1 = 0$, suggesting that \emph{the FT fine-tuning strategy should be adopted}. 

We extend the analysis of the previous section to the Gaussian mechanism \cite{abadi2016deep}, where gradients are $l_2$-bounded by $\xi_2$, and noise $\tilde{\mathbf{w}}_t^i \sim \mathcal{N}(0,\sigma_i^2\mathbb{I}_n)$, $\hat{\mathbf{w}}_t^i \sim \mathcal{N}(0,\sigma_i^2\mathbb{I}{p+n})$ are added with variance $\sigma_i^2 \ge \tfrac{c_2^2 \xi_2^2T}{d_i^2 \epsilon_i^2}\log \tfrac{1}{\delta_i}$.

\begin{theorem}
\label{theorem: gaussian-convergence}
    Let Assumptions \ref{asm: l-smooth} and \ref{asm: bounding-variances} hold. Let the learning rate $\eta = \frac{\beta}{\sqrt{T}}$ where $\beta < \frac{\sqrt{T}}{2L}$. The convergence rate of Pretrain-DPFL under the Gaussian Mechanism is as follows:
    \begin{equation}
        \label{eqs: gaussian-convergence}
        \begin{aligned}
            & \quad \mathbb{E}[\| \nabla F(\mathbf{x}_T) \|_2^2] \\
            &= \frac{1}{T} (\sum_{t=0}^{T_1} \mathbb E[\| \nabla F (\tilde{\theta}_t)\|_2^2] + \sum_{t=T_1 + 1}^{T} \mathbb{E} [\| \nabla F(\hat{\theta}_t) \|_2^2]) \\
            & \leq E_0 + E_1T_1 + E_2(T-T_1),
        \end{aligned}
    \end{equation}
    where $E_0 = \frac{4\{\mathbb{E}[F(\theta_{0})] - F^*\}}{\beta T^{1/2}} + \frac{8 \beta^2 L^2 c_2^2n\xi_2^2}{d^2}\sum_{i=1}^{N}\frac{1}{\epsilon_i^2}\log \frac{1}{\delta_i}$, $E_1 = \frac{4\beta}{T^{3/2}}(\frac{L\Lambda_1^2}{N} + 2L \Gamma) + \frac{16\beta^2L^2}{T^2}(G_1^2 + G_2^2)$, and $E_2 = \frac{4\beta}{T^{3/2}}(\frac{L\Lambda_2^2}{N} + 2L \Gamma) + \frac{8\beta^2L^2}{T^2}(\frac{c_2^2p\xi_2^2T}{d^2}\sum_{i=1}^{N}\frac{1}{\epsilon_i^2}\log \frac{1}{\delta_i})$.
\end{theorem}

Similarly, Theorem \ref{theorem: gaussian-convergence} shows that when $E_1 \geq E_2$, FT outperforms HT, whereas when $E_1 < E_2$, HT performs better.

\begin{figure}[t]
\begin{minipage}[b]{1.0\linewidth}
  \centering
  \centerline{\includegraphics[width=\linewidth]{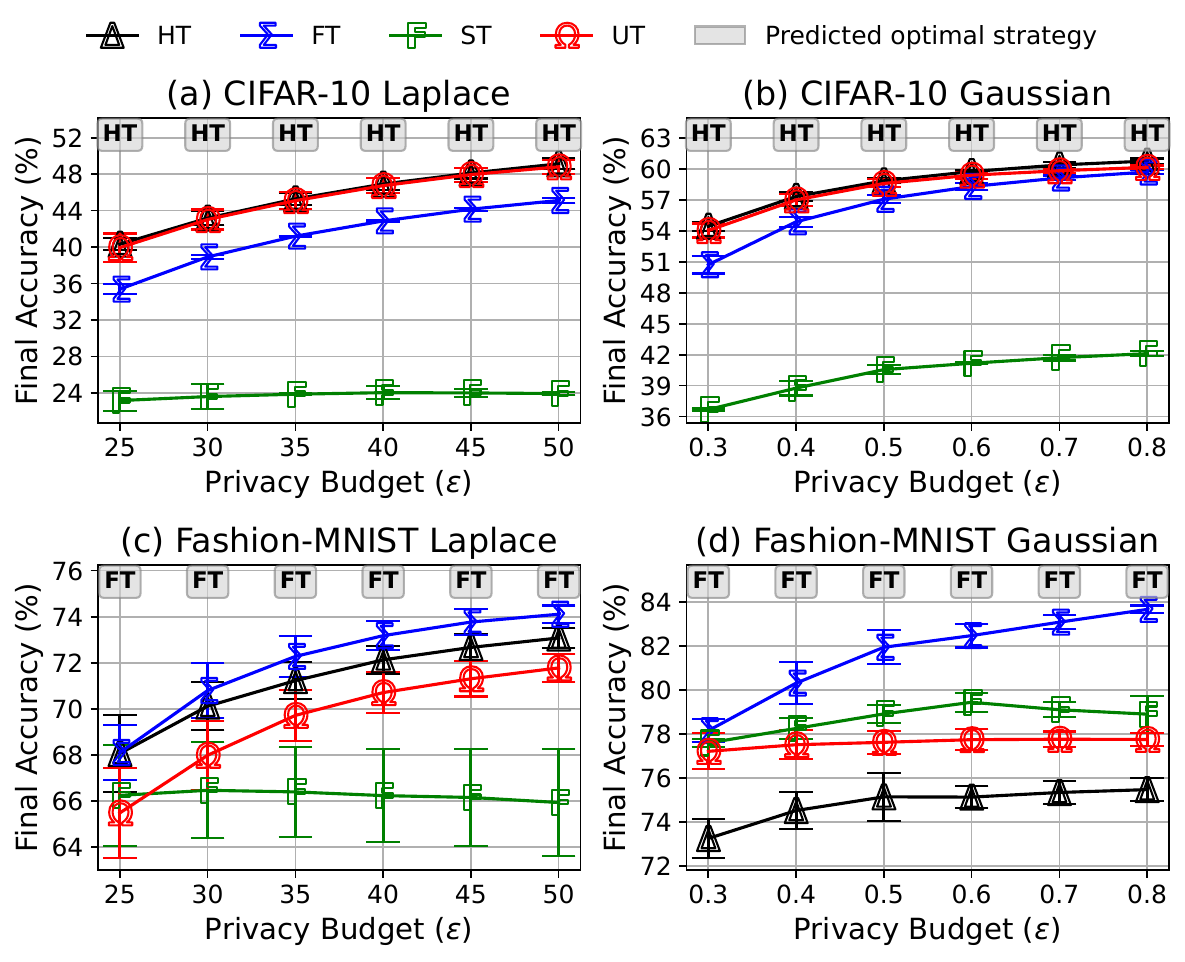}}
\end{minipage}
\vspace{-0.8cm}
\caption{Comparison of model accuracy across different privacy budgets using a CNN model, with the number of global training iterations set to $T = 128$.}
\vspace{-0.4cm}
\label{fig:laplace_gaussian_cnn}
\end{figure}

\begin{figure}[t]
\begin{minipage}[b]{1.0\linewidth}
  \centering
  \centerline{\includegraphics[width=\linewidth]{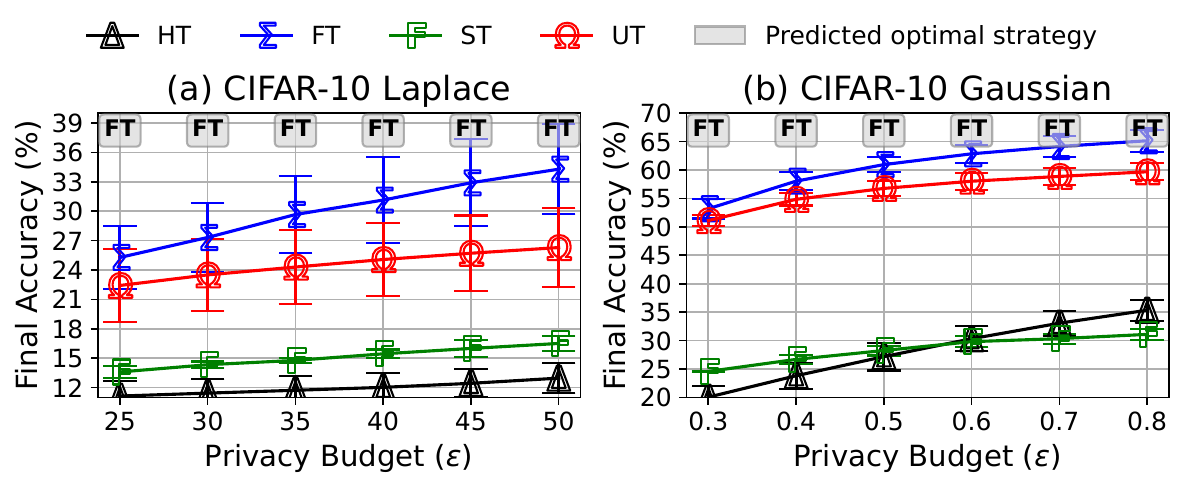}}
\end{minipage}
\vspace{-0.8cm}
\caption{Comparison of model accuracy across different privacy budgets using a ResNet20 model, with the number of global training iterations set to $T = 128$.}
\vspace{-0.4cm}
\label{fig:laplace_gaussian_resnet}
\end{figure}

\begin{figure}
\begin{minipage}[b]{1.0\linewidth}
  \centering
  \centerline{\includegraphics[width=\linewidth]{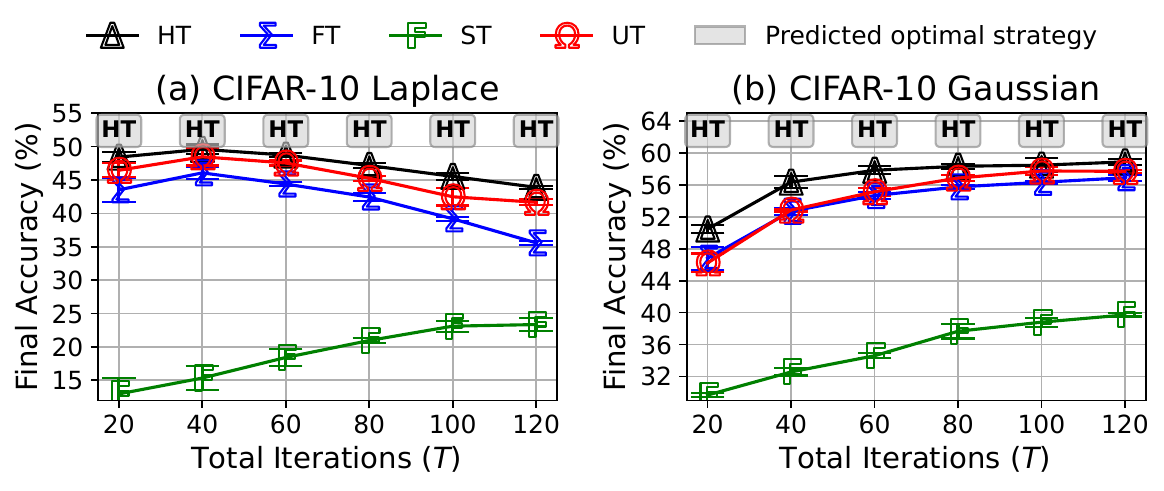}}
\end{minipage}
\vspace{-0.8cm}
\caption{Comparison of the model accuracy across different total iterations using a CNN model, with the $\epsilon$ set to 30 and 0.5 for the Laplace and Gaussian mechanisms, respectively.}
\vspace{-0.4cm}
\label{fig:laplace_gaussian_iter}
\end{figure}

\section{Experiments}

\subsection{Experiment Setup}

We pretrain on ImageNet-1K~\cite{imagenet} and fine-tune on CIFAR-10~\cite{zhou2023optimizing} (60k color images of size $3{\times}32{\times}32$) and Fashion-MNIST~\cite{zhou2023optimizing} (70k grayscale images of size $28{\times}28$). To match the pre-trained backbone, all images are resized to $3{\times}32{\times}32$. Considering data heterogeneity in FL, we partition data across $N{=}10$ clients using a Dirichlet sampler with $\alpha{=}1.0$~\cite{Hsu_Qi_Brown_2019}. 

We evaluate two models: a CNN with two convolutional and two fully connected layers~\cite{ma2023optimized} ($p{=}543{,}808$, $n{=}1{,}290$), and ResNet20~\cite{he2016deep} ($p{=}1{,}092{,}816$, $n{=}1{,}290$). The feature extractor is initialized from the ImageNet-1K checkpoint, while the classifier head is initialized using Kaiming initialization~\cite{he2015delving}. Four fine-tuning strategies are considered: HT, FT, and UT, together with ST (training from scratch) to isolate the effect of pretrained parameters. For UT, we set $T_1 \in \{0.25T, 0.5T, 0.75T\}$ and report results using the best-performing value.

Unless otherwise specified, the total number of global rounds is set to $T{=}128$. For the Laplace mechanism, gradient clipping bounds are $\xi_1{=}300$ (CNN) and $50$ (ResNet20); for the Gaussian mechanism, the bounds are $\xi_2{=}15$ (CNN) and $5$ (ResNet20) \cite{zhou2023optimizing}. The privacy budget is assumed identical across clients, i.e., $\epsilon_i{=}\epsilon$ \cite{ma2023optimized, zhou2023optimizing}, and the initial learning rate is selected by grid search. Each experiment is repeated three times, and the mean accuracy with error bars is reported.

Finally, according to Theorems~\ref{theorem: laplace-convergence} and~\ref{theorem: gaussian-convergence}, the relative performance of HT and FT depends on the magnitudes of $C_1$ and $C_2$ (or $E_1$ and $E_2$), which are influenced by $\epsilon$, model architecture, and total iterations $T$. The subsequent experiments are designed to examine these factors in detail.

\subsection{Experiment Results}
\emph{Effect of privacy budget $\epsilon$.} 
Fig.~\ref{fig:laplace_gaussian_cnn} shows CNN results under Laplace noise ($\delta{=}0$, $\epsilon{\in}[25,50]$) and Gaussian noise ($\delta{=}10^{-5}$, $\epsilon{\in}[0.3,0.8]$).
(i) The optimal strategies predicted by Pretrain-DPFL are consistent with the empirical winners. For example, under Gaussian noise with $\epsilon{=}0.8$, FT achieves 83.66\% on Fashion-MNIST, outperforming the second-best strategy by 8.19\%.
(ii) Pretrained fine-tuning methods consistently outperform ST. For instance, under Laplace noise with $\epsilon{=}50$, HT improves over ST by 25.22\% on CIFAR-10.

\emph{Impact of architecture.} Fig.~\ref{fig:laplace_gaussian_resnet} reports ResNet20 under the same settings. FT generally becomes stronger relative to HT because larger capacity reduces the relative impact of DP noise, which is consistent with the theory analysis.

\emph{Effect of total rounds $T$.} Varying $T{\in}\{20,\dots,120\}$ (Fig.~\ref{fig:laplace_gaussian_iter}) reveals a utility–privacy trade-off: small $T$ under-utilizes learning, whereas large $T$ amplifies accumulated noise. The predicted strategy remains correct across $T$, while the best $T$ depends on the privacy mechanism and dataset.

\textbf{Takeaways.} (1) Pretraining markedly improves DPFL utility over ST. (2) The HT vs.\ FT choice follows the predicted noise–learning balance and varies with $\epsilon$, model, and $T$. (3) Our selection rule reliably identifies the empirically optimal strategy across settings.

\section{Conclusion and Future Work}
As data privacy concerns grow, preserving privacy during model training has gained significant attention. Although DPFL is effective in protecting privacy, its applicability is hindered by the accuracy loss caused by noise. In our pioneering work, we explore how to optimally leverage pre-trained models to mitigate noise influence in DPFL. Specifically, we propose the Pretrain-DPFL framework, which explores the conditions under which the optimal fine-tuning strategy in DPFL can be identified to maximize the advantages of pre-trained models.
The superiority of our design is supported not only by rigorous theoretical analysis but also by extensive experiments. The results show that Pretrain-DPFL improves model accuracy by up to 25.22\% compared to ST and 8.19\% over the state-of-the-art baseline, underscoring its effectiveness in enhancing DPFL. Future work will extend the Pretrain-DPFL framework to broader scenarios, including multi-modal FL, personalized FL, and beyond.

%\vfill
\pagebreak

% \vfill\pagebreak
\footnotesize
\bibliographystyle{IEEEbib}
\bibliography{strings,refs}

\end{document}